\documentclass[letterpaper, 10 pt, conference]{ieeeconf}  

\IEEEoverridecommandlockouts                              

\usepackage{amsmath} 
\usepackage{amssymb}  
\usepackage{adjustbox}
\usepackage{bm}  
\usepackage{graphics} 
\usepackage{epsfig} 
\usepackage{wrapfig}
\usepackage{subcaption}
\usepackage[font=small,labelfont=small]{caption}
\usepackage{hyperref}
\usepackage{color}
\usepackage[table]{xcolor}
\usepackage[ruled, linesnumbered, vlined]{algorithm2e}
\usepackage{sidecap}
\usepackage{tikz}
\usetikzlibrary{arrows.meta}
\usepackage[noadjust]{cite}  

\graphicspath{{Figures/}}

\title{\LARGE \bf
On the Design of Region-Avoiding Metrics \\for Collision-Safe Motion Generation on Riemannian Manifolds
}

\author{Holger Klein, No\'emie Jaquier, Andre Meixner, and Tamim Asfour
\thanks{This work was supported by the Carl Zeiss Foundation through the JuBot project.  The authors are with the Institute for Anthropomatics and Robotics, Karlsruhe Institute of Technology, Karlsruhe, Germany. Correspondence to: {\tt\small holgerhugoklein@gmail.com, \{noemie.jaquier, andre.meixner, asfour\}@kit.edu}. }%
}

\newcommand{\trsp}{\mathsf{T}}

\DeclareMathOperator*{\argmin}{argmin} 

\newcommand{\euclideanspace}{\mathbb{R}}
\newcommand{\manifold}{\mathcal{M}}
\newcommand{\configmanifold}{\mathcal{Q}}
\newcommand{\tangentspace}[1]{\mathcal{T}_{#1}\mathcal{M}}

\newcommand{\innerprod}[3]{\langle #2, #3 \rangle_{#1}}  
\newcommand{\norm}[2]{\| #2\|_{#1}}  
\newcommand{\jointposition}{\bm{q}}
\newcommand{\jointvelocity}{\dot{\bm{q}}}
\newcommand{\jointacceleration}{\ddot{\bm{q}}}

\newcommand{\metric}{\bm{G}}
\newcommand{\kineticmetric}{\bm{M}}
\newcommand{\avmet}{\metric_a}
\newcommand{\barr}{b}
\newcommand{\avoidanceregion}{\configmanifold_a}
\newcommand{\dirtoregion}{\bm{v}}
\newcommand{\jac}{\bm{J}}

\newcommand{\armarVI}{\mbox{ARMAR-6}\xspace}

\definecolor{darkyellow}{rgb}{0.86, 0.66, 0.}
\definecolor{darkerred}{rgb}{0.9, 0.08, 0.}
\definecolor{darkred}{rgb}{0.66, 0.08, 0.23}
\definecolor{darkorange}{rgb}{1., 0.65, 0.0}
\definecolor{darkviolet}{rgb}{.6, 0., 0.6}
\definecolor{skyblue}{rgb}{0., 0.65, 0.9}
\definecolor{steelblue}{rgb}{0.275, 0.501, 0.706}
\definecolor{gray}{rgb}{0.7, 0.7, 0.7}

\DeclareRobustCommand{\blackarrow}{\raisebox{0.2pt}{\tikz{\draw[-{Latex[length=1.5mm, width=1.5mm]}, black,solid,line width = 1.1pt](0,0) -- (3mm,0);}}}
\DeclareRobustCommand{\redarrow}{\raisebox{0.2pt}{\tikz{\draw[-{Latex[length=1.5mm, width=1.5mm]}, darkerred,solid,line width = 1.1pt](0,0) -- (3mm,0);}}}
\DeclareRobustCommand{\bluearrow}{\raisebox{0.2pt}{\tikz{\draw[-{Latex[length=1.5mm, width=1.5mm]}, blue,solid,line width = 1.1pt](0,0) -- (3mm,0);}}}
\DeclareRobustCommand{\blackcircle}{\tikz{\filldraw[color=black, fill=black, thick](0,0) circle (.1);}}
\DeclareRobustCommand{\blackline}{\raisebox{2pt}{\tikz{\draw[black,solid,line width = 1.5pt](0,0) -- (3mm,0);}}}

\DeclareRobustCommand{\redline}{\raisebox{2pt}{\tikz{\draw[darkred,solid,line width = 1.5pt](0,0) -- (3mm,0);}}}
\DeclareRobustCommand{\orangeline}{\raisebox{2pt}{\tikz{\draw[darkorange,solid,line width = 1.5pt](0,0) -- (3mm,0);}}}
\DeclareRobustCommand{\bluedashedline}{\raisebox{2pt}{\tikz{\draw[steelblue,dashed,line width = 1.5pt](0,0) -- (3mm,0);}}}
\DeclareRobustCommand{\violetdashedline}{\raisebox{2pt}{\tikz{\draw[darkviolet,dashed,line width = 1.5pt](0,0) -- (3mm,0);}}}
\DeclareRobustCommand{\blueline}{\raisebox{2pt}{\tikz{\draw[steelblue,solid,line width = 1.5pt](0,0) -- (3mm,0);}}}
\DeclareRobustCommand{\violetline}{\raisebox{2pt}{\tikz{\draw[darkviolet,solid,line width = 1.5pt](0,0) -- (3mm,0);}}}

\begin{document}

\maketitle
\thispagestyle{empty}
\pagestyle{empty}

\begin{abstract}
The generation of energy-efficient and dynamic-aware robot motions that satisfy constraints such as joint limits, self-collisions, and collisions with the environment remains a challenge. In this context, Riemannian geometry offers promising solutions by identifying robot motions with geodesics on the so-called configuration space manifold. While this manifold naturally considers the intrinsic robot dynamics, constraints such as joint limits, self-collisions, and collisions with the environment remain overlooked. In this paper, we propose a modification of the Riemannian metric of the configuration space manifold allowing for the generation of robot motions as geodesics that efficiently avoid given regions. We introduce a class of Riemannian metrics based on barrier functions that guarantee strict region avoidance by systematically generating accelerations away from no-go regions in joint and task space. We evaluate the proposed Riemannian metric to generate energy-efficient, dynamic-aware, and collision-free motions of a humanoid robot as geodesics and sequences thereof.
\end{abstract}

\section{INTRODUCTION}
\label{sec:Introduction}
Robots must be able to generate motions that satisfy important constraints such as joint limits, self-collisions and collisions with obstacles in the environment. To do so, various approaches based on inverse kinematics (IK)~\cite{Chan95:JointLimitsAvoidanceIK,Maciejewski85:ObstacleAvoidanceIK,Chiacchio91:IKschemes,Rakita18:RelaxedIK,Rakita21:CollisionIK}, or motion planning~\cite{Elbanhawi14:SamplingBasedMotionPlanningReview, Gaebert21:MotionPlanningHumanLikeMotions,Zucker13:CHOMP,Kalakrishnan11:STOMP} have been proposed in the literature. 
Alternatively, the motion generation problem can be viewed through the lens of geometry by identifying robot motions with \emph{geodesics}, i.e., minimum-energy shortest paths, on Riemannian manifolds. In particular, the configuration space of a robot can be understood as a Riemannian manifold whose geometric properties are entirely defined by the robot dynamics via the so-called kinetic energy metric~\cite{Bullo05:GeometricControl}. In this context, geodesic trajectories account for the non-linearities arising from the intrinsic robot dynamics and can be leveraged to generate coherent, energy-optimal motions with respect to these dynamics~\cite{Jaquier2022:ISRR}. The configuration space manifold was also used to model human motions by conjecturing that point-to-point movements follow geodesics~\cite{Biess07:ComputationalModelPointing}, while complex movements result from sequences~\cite{Klein2022} and compositions~\cite{Neilson15:GeodesicSynergyHypothesis} thereof. Moreover, Sekimoto et al.~\cite{Sekimoto08:RiemannianDistanceBasedMotionPlanning} showed that geodesic motions of robots with human-like physical parameters resemble human motions. In our previous work~\cite{Klein2022}, we proposed a motion transfer framework that segments human arm motions into sequences of geodesics and reproduces them as geodesics in the robot configuration space. Each transferred geodesic is a minimum-energy trajectory accounting for the robot’s own inertial properties, while the overall sequence conserves the main characteristics of the original human motion.

While the kinetic energy metric disregards the influence of external forces, the configuration space of robots evolving in a potential field can be viewed as a Riemannian manifold endowed with the Jacobi metric~\cite{Casetti00:JacobiMetric}. Robot motions corresponding to geodesics on this manifold are optimal with respect to the total energy of the system. This representation was notably leveraged to study periodic trajectories of conservative mechanical systems~\cite{AlbuSchaffer22:ISRR} and to design controllers that select and transition between such trajectories~\cite{Sachtler22:StrictModes}.
Overall, viewing robot configuration spaces as Riemannian manifolds opens the door to the generation of highly-efficient motions by considering the intrinsic robot dynamics. However, the obtained trajectories do not consider joint limits, self-collision, or obstacle avoidance.

In this paper, we propose to modify the configuration space manifold so that geodesics generate collision-free trajectories within joint limits, while remaining energy-optimal. This is achieved by reshaping the underlying Riemannian metric around no-go regions --- i.e., robot configurations beyond joint limits or resulting in collisions --- so that the energy required to pass through these regions becomes prohibitively high. To do so, we design a class of region-avoiding metrics based on barrier functions and provide a necessary and sufficient condition for guaranteed strict region avoidance (Sections~\ref{subsec:BarrierFunctions} and~\ref{subsec:RegionAvoidingMetrics}). Such metrics generate geodesic accelerations generally pointing away from no-go regions. To improve their efficiency, we propose a change of basis resulting in region-avoiding metrics that generate direction-dependent geodesic accelerations (Section~\ref{subsec:BasisChange}). Namely, the repelling acceleration is maximal, respectively minimal, when the robot moves towards, respectively along, no-go regions. While joint limit avoidance metrics are naturally defined in joint space, self- and obstacle avoidance metrics are easier to define in task space. To compute geodesics in the configuration space manifold, we propose to pullback directions to the obstacle from task space, so that the resulting joint-space metric leads to the same direction-dependent acceleration in task space (Section~\ref{subsec:TaskSpaceAvoidance}). Finally, we combine the configuration space metric with regions-avoiding metrics (Section~\ref{subsec:CombiningMetrics}) and generate energy-efficient, dynamic-aware, and collision-free robot motions by following geodesics in the corresponding manifold (Section~\ref{sec:GeodesicMotionGeneration}).

The contributions of this paper are threefold: \emph{(i)} We propose a class of direction-dependent Riemannian metrics for strict region-avoidance; \emph{(ii)} We provide pullback operations that conserve the properties of such region-avoiding metrics; and \emph{(iii)} We generate energy-efficient, dynamic-aware robot motions as collision-free geodesics in a modified configuration space manifold. We evaluate the ability of our approach to stay within joint limits and to avoid self-collisions and obstacles in various scenarios (Section~\ref{sec:Experiments}). 
A video of the experiments accompanies the paper and is available at \url{https://youtu.be/qT43XgYOlU0}.

\section{RELATED WORK}
\label{sec:RelatedWork}
From early on, kinematic redundancy has been exploited to introduce additional feasibility criteria such as joint limits or obstacle avoidance in IK frameworks. This was achieved by computing weighted least norm IK solutions~\cite{Chan95:JointLimitsAvoidanceIK} or by resorting to nullspace-based task augmentation and task priority strategies~\cite{Maciejewski85:ObstacleAvoidanceIK, Chiacchio91:IKschemes}. Recent works~\cite{Rakita18:RelaxedIK,Rakita21:CollisionIK} incorporated various feasibility criteria --- including motion smoothness, self- and obstacle collision avoidance --- into IK solvers by formulating IK as a multi-objective non-linear optimization problem. In contrast to IK, motion planning algorithms are explicitly concerned about finding a collision-free trajectory between boundary points~\cite{Elbanhawi14:SamplingBasedMotionPlanningReview}. Such algorithms are generally guaranteed to find a solution if one exists, and can be augmented with an optimization phase to satisfy additional criteria~\cite{Gaebert21:MotionPlanningHumanLikeMotions}. Other motion planning algorithms~\cite{Zucker13:CHOMP,Kalakrishnan11:STOMP} directly optimize entire trajectories by minimizing multi-objective costs including an obstacle avoidance term. However, IK and motion planning algorithms are not straightforwardly applicable on Riemannian manifolds with non-constant metric.

Several works proposed to generate robot motions as geodesics on a Riemannian manifold. Obstacle avoidance was achieved in~\cite{Mainprice16:RieMO,Laux21:GeodesicMotionPlanning} by reshaping the identity metric of the Euclidean end-effector space around obstacles, so that they are naturally avoided by geodesics. Although we follow a similar idea, our approach ensures that the behavior of the robot is generally characterized by the configuration space metric. Ratliff et al.~\cite{Ratliff18:RMPs} proposed to combine local Riemannian policies encoding different robot behaviors. In this case, the Riemannian metrics quantify the directional importance of each local policy, which can be combined in a common space via pullback operations~\cite{Cheng21:RMPflow}. 
Our approach differs as we focus on reshaping a given metric, i.e., the configuration space metric, to incorporate region avoidance terms. Robot trajectories are then entirely determined by the reshaped metric.

While the aforementioned approaches define the Riemannian metric of interest manually, robot motions have also been generated on learned stochastic Riemannian manifolds. Such manifolds are obtained by endowing the latent space of a latent variable model with a Riemannian metric reflecting the support of the training data~\cite{Tosi14:RiemannianGPLVM,Arvanitidis18:LatentSpaceOddity} and geodesics naturally avoid regions far from the data. Scannell et al.~\cite{Scannell21:TrajectoryOptLearnedManifold} learned a Riemannian manifold reflecting the spatial distribution of transition dynamic modes of a controlled quadcopter. Regions with turbulent modes are then avoided as a soft constraint by following geodesics on the learned manifold.
Beik-Mohammadi et al.~\cite{BeikMohammadi22:ReactiveMotionGeneration} proposed to represent robotic skills in a latent Riemannian manifold learned from demonstrations. Trajectories resembling the demonstrations are then produced as geodesics on the learned manifold. Following~\cite{Arvanitidis21:GeometricallyEnrichedLatentSpaces}, the authors reshaped the learned metric to avoid obstacles by pulling back a task-space region-avoiding metric based on the exponential barrier function. However, this metric results in a soft constraint and does not guarantee collision avoidance. In contrast, we introduce Riemannian metrics resulting in strict region avoidance, whose properties are preserved by pullback operations.

\section{BACKGROUND}
\label{sec:Background}
In this section, we introduce the mathematical tools needed to generate robot motions as geodesics on Riemannian manifolds. 
We refer the interested reader to, e.g.,~\cite{DoCarmo92:RiemannianGeometry,Lee18:RiemannianManifolds}, and to~\cite{Bullo05:GeometricControl} for in-depth introductions to Riemannian geometry, and geometry of mechanical systems, respectively.

\subsection{Riemannian Manifolds and Riemannian Metrics}
\label{subsec:RiemannianManifolds}
A $d$-dimensional manifold $\manifold$ is a topological space, which is locally Euclidean. In other words, each point in $\manifold$ has a neighborhood that is homeomorphic to an open subset
of the $d$-dimensional Euclidean space $\euclideanspace^d$, also called a chart. The manifold $\manifold$ is smooth if differentiable transitions between charts can be defined. 
A tangent space $\tangentspace{\jointposition}$ is associated with each point $\jointposition\in\manifold$ and is formed by the differentials at $\jointposition$ to all curves on $\manifold$ passing through $\jointposition$. The disjoint union of all tangent spaces $\tangentspace{\jointposition}$ forms the tangent bundle $\tangentspace{}$.
A Riemannian manifold is a smooth manifold equipped with a Riemannian metric, i.e., a smoothly-varying inner product acting on $\tangentspace{}$. Given a choice of local coordinates, the Riemannian metric is represented as a symmetric positive-definite matrix $\metric(\jointposition)$, which depends smoothly on $\jointposition\in\manifold$.
The Riemannian metric leads to local, nonlinear expressions of inner products and angles.
Specifically, the Riemannian inner product between two velocity vectors $\bm{u}$, $\bm{v}\in\tangentspace{\jointposition}$ at $\jointposition\in\manifold$ is given as
\begin{equation}
    \innerprod{\jointposition}{\bm{u}}{\bm{v}} = \langle \bm{u}, \metric(\jointposition)\bm{v} \rangle
     \;= \bm{u}^\trsp \metric(\jointposition)\bm{v}.
    \label{Eq:InnerProduct}
\end{equation}
The Riemannian norm is defined as $\norm{\metric(\jointposition)}{\bm{v}} = \sqrt{\innerprod{\jointposition}{\bm{v}}{\bm{v}}}$. 
Importantly, the Riemannian metric fully captures the geometry of the manifold. Therefore, it also allows us to compute shortest paths on $\manifold$, as explained next.

\subsection{Geodesics}
\label{subsec:Geodesics}
Similarly to straight lines in Euclidean space, geodesics are minimum-energy and minimum-length curves on Riemannian manifolds. They follow from the application of the Euler-Lagrange equations to the kinetic energy defined by the Riemannian metric as
\begin{equation}
    k = \int \frac{1}{2} \norm{\metric(\jointposition(t))}{\jointvelocity(t)}^2 \mathrm{d}t. 
    \label{Eq:KineticEnergy}
\end{equation}
Geodesics solve the following system of second-order ordinary differential equations (ODE)\footnote{For brevity, we omit the dependency of $\jointposition$ on $t$ when obvious.}
\begin{equation}
    \ddot{q}_i + \sum_{jk} \Gamma_{jk}^i \dot{q}_j \dot{q}_k = 0,
    \label{Eq:Geodesic}
\end{equation}
with $\Gamma_{jk}^i$the Christoffel symbols of the second kind given by
\begin{equation*}
    \Gamma_{jk}^i = \frac{1}{2} \sum_l g^{-1}_{il} \left( \frac{\partial g_{lj}}{\partial q_k} + \frac{\partial g_{lk}}{\partial q_j} - \frac{\partial g_{jk}}{\partial q_l} \right), 
\end{equation*}
and $q_k$ and $g_{ij}$ denoting the $k$-th and $i,j$-th component of $\jointposition$ and $\metric$. 
In other words, geodesic trajectories are obtained by applying the joint acceleration $\jointacceleration(t)$ solution of Eq.~\eqref{Eq:Geodesic} at each configuration $\jointposition(t)$ with velocity $\jointvelocity(t)$ along the trajectory. Notice that the velocity norm is constant along a geodesic, i.e, $\norm{\metric(\jointposition(t))}{\jointvelocity(t)}=c$.

\subsection{The Configuration Space Manifold}
The configuration space $\configmanifold$ of a robot can be viewed as a smooth manifold with a simple global chart. Points on this manifold correspond to different joint configurations $\jointposition\in\configmanifold$. 
The configuration manifold $\configmanifold$ of mechanical systems can be endowed with the kinetic-energy metric~\cite{Bullo05:GeometricControl} or with the Jacobi metric~\cite{Casetti00:JacobiMetric}. The kinetic-energy metric is equal to the robot mass-inertia matrix $\kineticmetric(\jointposition)$, while the Jacobi metric additionally incorporates external potentials. Intuitively, both metrics curves the space so that the configuration manifold accounts for the robot nonlinear inertial properties. In the case of the configuration space manifold, the geodesic equation~\eqref{Eq:Geodesic} corresponds to the standard equation of motion 
\begin{equation}
     \kineticmetric(\jointposition)\jointacceleration + \bm{C}(\jointposition, \jointvelocity)\jointvelocity + U(\jointposition) = 0
\end{equation}
with $U(\jointposition)=0$ for the kinetic-energy metric. Therefore, geodesics correspond to passive trajectories of the system and thus intrinsically accounts for its dynamic properties. 

\section{RIEMANNIAN METRICS FOR STRICT REGION AVOIDANCE}
\label{sec:RiemannianMetricsRegionAvoidance}
In the following, we assume without loss of generality that the configuration manifold is endowed with the kinetic-energy metric $\kineticmetric$. Robot motions generated by geodesics are energy-optimal with respect to the robot dynamics. However, they overlook joint limits and do not avoid self-collisions and obstacles in the workspace. Thus, we introduce a class of Riemannian metrics for strict region avoidance and use them to reshape the metric $\kineticmetric$ around no-go regions and obtain a collision-free metric $\metric$.

\subsection{The 1-D case: Barrier Functions For Region Avoidance}
\label{subsec:BarrierFunctions}
We define the collision-free metric $\metric$ as a combination of the kinetic-energy metric $\kineticmetric$ with region-avoiding metrics $\avmet$. Intuitively, region-avoiding metrics are designed so that geodesics $\jointposition(t) \in \configmanifold$ passing through the corresponding region $\avoidanceregion \subset \configmanifold$ have prohibitively high length. Moreover, the metric $\avmet$ should not influence the geodesics away from the avoided region. 
Therefore, for strict region avoidance, the collision-free metric $\metric(\jointposition)$ should satisfy
\begin{equation}
\norm{\metric(\jointposition)}{\jointvelocity} =    
\begin{cases} 
      \infty & \text{ if } \jointposition \subset \avoidanceregion, \\
      \norm{\kineticmetric(\jointposition)}{\jointvelocity} & \text{ otherwise.} \\
    \end{cases}
    \label{Eq:MetricRequirements}
\end{equation}
In this paper, we propose to build region-avoiding metrics based on the well-studied family of \textit{barrier functions}~\cite{Boyd04:ConvexOpt}. Barrier functions $\barr : \configmanifold \to \euclideanspace$ become very large near a given point $\jointposition_a$ and vanish everywhere else, thus may satisfy --- in the 1-D case --- the requirements for $\avmet$. 
Notice that, due to the aforementioned properties, barrier functions are also often used as avoidance costs when optimizing robot trajectories~\cite{Rakita18:RelaxedIK,Rakita21:CollisionIK}.

Here, we consider the following three functions:
\begin{itemize}
    \item the exponential barrier, $\barr_{\text{exp}}(\jointposition) = \sigma \exp\big(-\frac{(\jointposition-\jointposition_a)^2}{\lambda^2}\big)$,
    \item the logarithmic barrier, $\barr_{\text{log}}(\jointposition) = -\sigma\log(\jointposition-\jointposition_a)$,
    \item the inverse barrier, $\barr_{\text{inv}}(\jointposition) = \frac{\sigma}{\jointposition-\jointposition_a}$,
\end{itemize}
where $\sigma$ is a scaling factor and $\lambda$ determines the width of the exponential barrier.
To reason about the effect of the different barrier functions on geodesics, we first assume a 1-D configuration space manifold $\configmanifold \equiv \euclideanspace$ with metric $m(q)$ and a point $q_a$ to avoid. In this case, we can construct a simple metric $g(q) = m(q) + \barr(q)$ resulting in the norm
\begin{equation}
    \norm{g(q)}{\dot{q}} = \sqrt{\big(m(q) + \barr(q)\big)\dot{q}^2}.
    \label{Eq:NormBarrierMetric}
\end{equation}
Away from $q_a$, the barrier function vanishes, i.e., $b(q)\to 0$, and $\norm{g(q)}{\dot{q}} =\norm{m(q)}{\dot{q}}$. This means that geodesics are entirely characterized by $m(q)$ away from $q_a$, i.e., $g(q)$ fulfills the second requirement of Eq.~\ref{Eq:MetricRequirements}. The first requirement, i.e., $\norm{g(q)}{\dot{q}} \to \infty$ if $q=q_a$, is satisfied for all velocities $\dot{q}$ if and only if $\barr(q)\to\infty$ when $q = q_a$.
Notice that we can observe the effect of such a metric on geodesics. Since the velocity norm is constant along a geodesic, i.e., $\norm{g(q)}{\dot{q}}=c$ (see Section~\ref{subsec:Geodesics}), the geodesic velocity is given as
\begin{equation}
    |\dot{q}| = \frac{c}{\sqrt{\big(m(q) + \barr(q)\big)}},
\end{equation}
using Eq.~\eqref{Eq:NormBarrierMetric}.
Therefore, when $\barr(q)\to\infty$, the velocity is $\dot{q}=0$ and the geodesic cannot cross the no-go region. 
In other words, \emph{strict region avoidance is guaranteed if and only if the barrier function goes to infinity} in the region that should be avoided. Therefore, the inverse and logarithmic barrier results in hard collision-avoidance constraints. Although the exponential barrier increases the length of geodesics around no-go regions, it only results in a soft constraint. This implies that geodesics with high-enough velocity norm can cross no-go regions when using the exponential barrier.

To gain additional insights about geodesics generated by $g(q)$, we study the geodesic acceleration around no-go regions. For simplicity and without loss of generality, we assume a constant metric $m(q)=1$ and a no-go region ${q_a = 0}$.
In one dimension, the geodesic equation~\eqref{Eq:Geodesic} simplifies as
\begin{equation}
    \ddot{q} = -\Gamma \dot{q}^2 \;\;\;\text{ with }\;\;\; \Gamma = \frac{1}{2} \frac{\barr'(q)}{1+\barr(q)},
    \label{eq:geodesic_acceleration_1d}
\end{equation}
and $\barr'(q) = \frac{\partial \barr(q)}{\partial q}$.
This shows that the geodesic acceleration does not only depends on the metric, but rather on the relationship between the metric and its gradient. The geodesic accelerations obtained from $g(q)$ with different barrier functions are
\begin{align}
    \ddot{q}_{\text{exp}} &= \frac{q}{\frac{\lambda^2}{\sigma}\exp\big(\frac{q^2}{\lambda^2}\big) + \lambda^2} \; \dot{q}^2, \\
    \ddot{q}_{\text{log}} &= \frac{1}{2}\frac{\sigma}{q-\sigma q\log(q)} \; \dot{q}^2, \\
    \ddot{q}_{\text{inv}} &=\frac{1}{2q} \; \dot{q}^2.
\end{align}
We observe that the metrics based on inverse and logarithmic barriers generate infinite accelerations for all velocities when $q\to q_a = 0$. Therefore, the geodesics approaching $q_a$ not only have large lengths but also accelerate away from the barrier. Instead, the exponential barrier generates no acceleration when $q=q_a=0$ and geodesics may pass through no-go regions.
Notice that the metric based on the logarithmic barrier also produces infinite accelerations when ${\sigma\log(q)=1}$, i.e., it also influences geodesics around $q = \log^{-1}(\sigma^{-1})$. 
Therefore, the inverse barrier is the most suited to build 1-D region-avoiding Riemannian metrics satisfying Eq.~\eqref{Eq:MetricRequirements}. Powers of this barrier, i.e., $\barr(\jointposition) = \frac{\sigma}{(\jointposition-\jointposition_a)^n}$, have similar properties and may also be used to build such metrics. 

\subsection{From Barrier Functions to Region-Avoiding Metrics}
\label{subsec:RegionAvoidingMetrics}
After discussing 1-dimensional region-avoiding metrics based on barrier functions, we aim at generalizing such metrics to avoid no-go regions in $d$-dimensional spaces $\configmanifold$.  We define $\dirtoregion=\jointposition-\jointposition_a$ as the vector pointing from the current joint angles $\jointposition$ to the closest point $\jointposition_a\in\avoidanceregion$.
Barrier functions are then computed using the Euclidean norm of $\dirtoregion$ as $\barr(\norm{}{\dirtoregion})$
and a collision-free metric $\metric(\jointposition)$ can be constructed as
\begin{equation}
    \metric(\jointposition) = \kineticmetric(\jointposition) + \metric_a(\jointposition) \;\;\;\text{with}\;\;\; \metric_a(\jointposition)=  \barr(\norm{}{\dirtoregion})\:\bm{I}_d,
    \label{Eq:CollisionFreeMetric}
\end{equation}
Following the same argument as in the $1$-D case, it can easily be shown that \emph{the metric $\metric(\jointposition)$ guarantees strict region avoidance if and only if the barrier function $b$ goes to infinity for $\jointposition\in\configmanifold_a$}. Note that a similar metric was proposed in~\cite{BeikMohammadi22:ReactiveMotionGeneration} based on the exponential barrier in Euclidean space, i.e., $\kineticmetric(\jointposition)=\bm{I}_d$. However, as already stated, the exponential barrier does not provide strict collision avoidance. 

\begin{figure}
    \centering
    \begin{subfigure}[b]{.3\linewidth}
        \includegraphics[trim={1cm 1.2cm 2.5cm 2.5cm},clip,width=\textwidth]{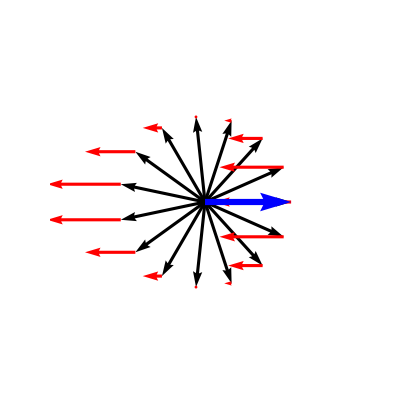}
        \caption{$\dirtoregion$ as basis vector}
        \label{fig:geod_accs_diagonal_v_base_vec}
    \end{subfigure}
    \begin{subfigure}[b]{.3\linewidth}
        \includegraphics[trim={1cm 1.2cm 2.5cm 2.5cm},clip,width=\textwidth]{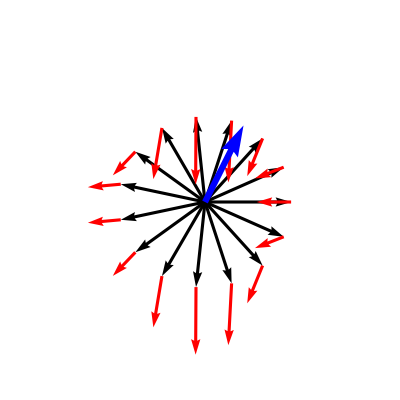}
        \caption{No basis change}
        \label{fig:geod_accs_diagonal}
    \end{subfigure}
    \begin{subfigure}[b]{.3\linewidth}
        \includegraphics[trim={1cm 1.2cm 2.5cm 2.5cm},clip,width=\textwidth]{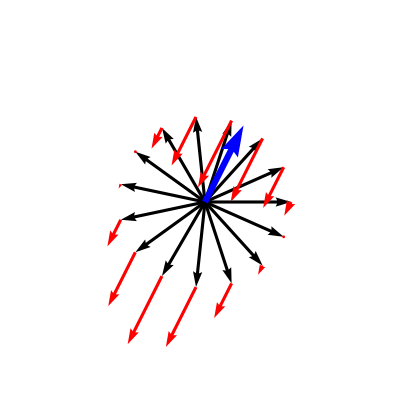}
        \caption{With basis change}
        \label{fig:geod_accs_base_change}
    \end{subfigure}
    \caption{Geodesic accelerations (\redarrow) generated by different region-avoiding metrics $\avmet$ in $\configmanifold$ and acting on planar geodesics with different velocities (\blackarrow). The vector $\dirtoregion$ (\bluearrow) indicates the direction from the current geodesic position to $\configmanifold_a$.  
    }
    \label{fig:geodesic_accs_diagonal_and_base_change_avoidance_metric}
    \vspace{-0.3cm}
\end{figure}

As previously, we are interested in the influence of the metric~\eqref{Eq:CollisionFreeMetric} on geodesics around the no-go region $\configmanifold_a$ and study the corresponding geodesic acceleration. For simplicity, we assume $\configmanifold\equiv \euclideanspace^d$ for which $\kineticmetric(\jointposition) = \bm{I}_d$. Figures~\ref{fig:geod_accs_diagonal_v_base_vec}-~\ref{fig:geod_accs_diagonal} display the geodesic acceleration produced by $\metric$ for geodesics approaching $\configmanifold_a$ from different directions $\dirtoregion$ and with different velocities $\jointvelocity$. We observe that, in the multidimensional case, the generated geodesic acceleration depends on the direction of $\jointvelocity$. Moreover, when $\dirtoregion$ is parallel to a canonical basis vector, e.g., $\dirtoregion=(1, 0)^\trsp$ in Figure~\ref{fig:geod_accs_diagonal_v_base_vec}, the geodesic acceleration points in the direction $-\dirtoregion$ with a magnitude proportional to $\innerprod{\jointposition}{\dirtoregion}{\jointvelocity}$. This is the optimal behavior as \emph{(i)} the acceleration is maximal when $\jointvelocity$ is aligned with $\dirtoregion$, i.e., for geodesics traveling towards $\configmanifold_a$, and \emph{(ii)} the acceleration is zero when $\jointvelocity$ is orthogonal to $\dirtoregion$, i.e., the metric does not influence geodesics traveling along the no-go region.
However, as shown in Figure~\ref{fig:geod_accs_diagonal}, this no longer applies when $\dirtoregion$ is not a basis vector, i.e., geodesics (including those traveling along $\avoidanceregion$) are irregularly accelerated away from $\avoidanceregion$. This issue is alleviated by changing the basis of the metric, as explained next.
 
\subsection{Efficient Region Avoidance via Basis Change}
\label{subsec:BasisChange}
To build region-avoiding metrics that systematically generate geodesic accelerations away from $\avoidanceregion$, we first study the geodesic equation~\eqref{Eq:Geodesic} in the case where $\dirtoregion$ is parallel to a canonical basis vector. 
Namely, as $\avmet$ is a diagonal matrix, its derivative is
\begin{equation*}
    \frac{\partial (g_a)_{ij}}{\partial q_l} =
    \begin{cases} 
      \frac{v_l \barr'(\norm{}{\dirtoregion})}{\norm{}{\dirtoregion}}  & \text{if } i = j, \\
      0 & \text{otherwise}. \\
    \end{cases}
\end{equation*}
Moreover, if $\dirtoregion$ is aligned with the $m$-th canonical basis vector, we have $\dirtoregion=(0, \ldots, v_m, \ldots, 0)^\trsp$, with $v_m$ the $m$-th component of $\dirtoregion$.
In this case, the derivative is 
\begin{equation*}
    \frac{\partial (g_a)_{ij}}{\partial q_l} =
    \begin{cases} 
      \barr'(\norm{}{\dirtoregion})=\barr'(v_m) & \text{if } i = j \text{ and } l = m, \\
      0 & \text{otherwise}. \\
    \end{cases}
\end{equation*}
From this it follows that the only non-zero component of the geodesic acceleration is  
\begin{equation}
    \ddot{q}_m =
        -\frac{1}{2}\frac{\barr'(v_m)}{1 + \barr(v_m)} \dot{q}_m^2,
\end{equation}
i.e., the geodesic acceleration points in the direction $-\dirtoregion$. Note that $\ddot{q}_m$ is identical to the geodesic acceleration computed in the 1-D case, see Eq.~\eqref{eq:geodesic_acceleration_1d}.
To generate the same behaviour for any direction $\dirtoregion$, we propose to define $\avmet$ via a change of basis. 
To do so, we first generate a linearly independent system of vectors $\{\dirtoregion, \bm{r_2},...,\bm{r_{d}}\}$, and use the Gram-Schmidt algorithm to turn this system into an orthonormal basis $\bm{B} = (\bm{b_1}, ...,\bm{b_d})$, where $\bm{b_1} =\dirtoregion/\norm{}{\dirtoregion}$. We then construct the metric $\avmet^{\bm{B}}$ in this basis following Eq.~\eqref{Eq:CollisionFreeMetric}. Finally, we transform the components of the metric into the canonical basis, so that $\avmet = \bm{B}\avmet^{\bm{B}}\bm{B}^{\trsp}$. Figure~\ref{fig:geod_accs_base_change} shows an example of the resulting geodesic accelerations.

\subsection{Region Avoidance via Pullback Operations}
\label{subsec:TaskSpaceAvoidance}
In the previous sections, we assumed identical domains $\configmanifold$ for the region-avoiding and kinetic-energy metrics. However, avoidance regions corresponding to self-collision and obstacles are naturally encoded as positions in the robot task space. Here, we aim at designing a metric $\metric$ generating geodesics $\jointposition(t) \subset \configmanifold$ which avoid collisions in task space. In this case, obstacles are defined as region in $\euclideanspace^3$, and $\dirtoregion_{\bm{x}} = \bm{x}-\bm{x}_a$ is a vector from the closest point $\bm{x}\in\euclideanspace^3$ on the robot's collision geometry to an obstacle $\bm{x_a}\in\euclideanspace^3$. 

\begin{figure}
    \centering
    \begin{subfigure}[b]{.45\linewidth}
        \includegraphics[trim={3cm 2cm 2.5cm 2.5cm},clip,width=.9\textwidth]{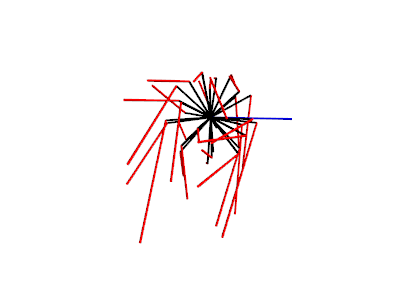}
        \caption{Pullback metric}
        \label{fig:geod_accs_pullback_metric}
    \end{subfigure}
    \begin{subfigure}[b]{.45\linewidth}
        \includegraphics[trim={3cm 2cm 2.5cm 2.5cm},clip,width=.9\textwidth]{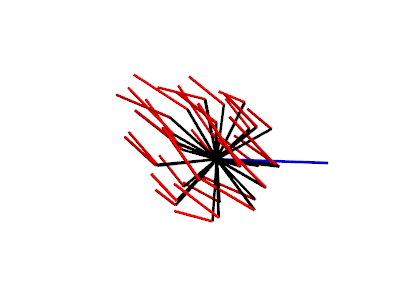}
        \caption{Pullback direction}
        \label{fig:geod_accs_pullback_direction}
    \end{subfigure}
    \caption{Geodesic accelerations generated by different pullback operations from $\euclideanspace^3$ to $\configmanifold$ and acting on geodesics with different velocities. The resulting accelerations $(\redarrow)$ and the velocities (\blackarrow) are represented in \emph{task space}. The vector $\dirtoregion$ (\bluearrow) indicates the direction from the current position to the obstacle in $\euclideanspace^3$.  
    }
    \label{fig:geodesic_accs_pullback}
    \vspace{-0.3cm}
\end{figure}

Region-avoiding metrics $\metric_a^{\bm{B}}(\bm{x}) = b(\norm{}{\dirtoregion_{\bm{x}}})\bm{I}_3$ can then be defined in $\euclideanspace^3$ and pulled back into the configuration space manifold $\configmanifold$ using the forward kinematic function $f$ similarly as in~\cite{Arvanitidis21:GeometricallyEnrichedLatentSpaces} and~\cite{BeikMohammadi22:ReactiveMotionGeneration}.
Namely, given a smooth function $f: \configmanifold \to \manifold$, a metric $\metric_\manifold$ on $\manifold$ is pulled back onto $\configmanifold$ as
\begin{equation}
    \metric_\configmanifold = \jac_f^\trsp \metric_\manifold \jac_f,
\end{equation}
with $\jac_f$ the Jacobian of $f$, and $\manifold\equiv\euclideanspace^3$ in our case. However, despite defining $\metric_a^{\bm{B}}(\bm{x})$ as in Section~\ref{subsec:BasisChange}, the resulting pullback metric does not generate geodesic accelerations that systematically point away from the obstacle in task space (see Figure~\ref{fig:geod_accs_pullback_metric}).
Therefore, we propose to instead pull back $\dirtoregion_{\bm{x}}\in\euclideanspace^3$ using the Jacobian pseudo inverse $\jac^\dagger$ as
\begin{equation}
    \dirtoregion = \jac^\dagger \dirtoregion_{\bm{x}},
    \label{Eq:Pulledback_direction}
\end{equation}
and to construct $\avmet$ directly in joint space using Eq.~\ref{Eq:CollisionFreeMetric}. This can be justified by the fact that, in theory, we could infer all
 $\jointposition_a\in\configmanifold$ for which a collision occurs and construct $\dirtoregion=\jointposition-\jointposition_a$ directly in $\configmanifold$. The pullback~\eqref{Eq:Pulledback_direction} can be seen as a low-cost estimation of $\dirtoregion$. As shown in Figure~\ref{fig:geod_accs_pullback_direction}, the resulting metric generates geodesic accelerations roughly aligned with $-\dirtoregion_{\bm{x}}$ in task space, which are maximal when the velocity $\dot{\bm{x}}=\jac\jointvelocity$ is aligned with $\dirtoregion_{\bm{x}}$ and zero when $\dot{\bm{x}}$ is orthogonal to  $\dirtoregion_{\bm{x}}$.

\subsection{Combining Riemannian Metrics}
\label{subsec:CombiningMetrics}
Using the region-avoiding metrics defined in Sections~\ref{subsec:RegionAvoidingMetrics}-~\ref{subsec:TaskSpaceAvoidance}, we define the overall metric $\metric$ as the sum 
\begin{equation}
    \metric(\jointposition) = \kineticmetric(\jointposition) + \metric_j(\jointposition) + \metric_s(\jointposition) + \metric_o(\jointposition),
    \label{Eq:SumOfMetrics}
\end{equation}
where $\metric_j$ is a diagonal metric with $g_{ii}=b(q_{i}^\ell)$ ensuring that geodesics stay within the joint limits $\{q_{i}^\ell\}_{i=1}^d$, while $\metric_s$ and $\metric_o$ are self-collision and obstacle avoidance metrics defined from the pullback direction in task space (see Section~\ref{subsec:TaskSpaceAvoidance}). As the resulting geodesics are energy-minimizing with respect to the sum of metrics, they combine their different requirements. Moreover, as the influence of our region-avoiding metrics vanishes away from joint limits and obstacles thanks to the properties of barrier functions, the geodesics are characterized by the kinetic-energy metric away from no-go regions. Therefore, the metric~\eqref{Eq:SumOfMetrics} generates geodesics leading to energy-efficient, dynamic-aware, and collision-free robot motions. The computation of such geodesics is detailed next.

\section{GEODESIC COMPUTATION}
\label{sec:GeodesicMotionGeneration}
\begin{figure}
    \centering
    \begin{subfigure}[b]{.45\linewidth}
        \includegraphics[width=.9\textwidth]{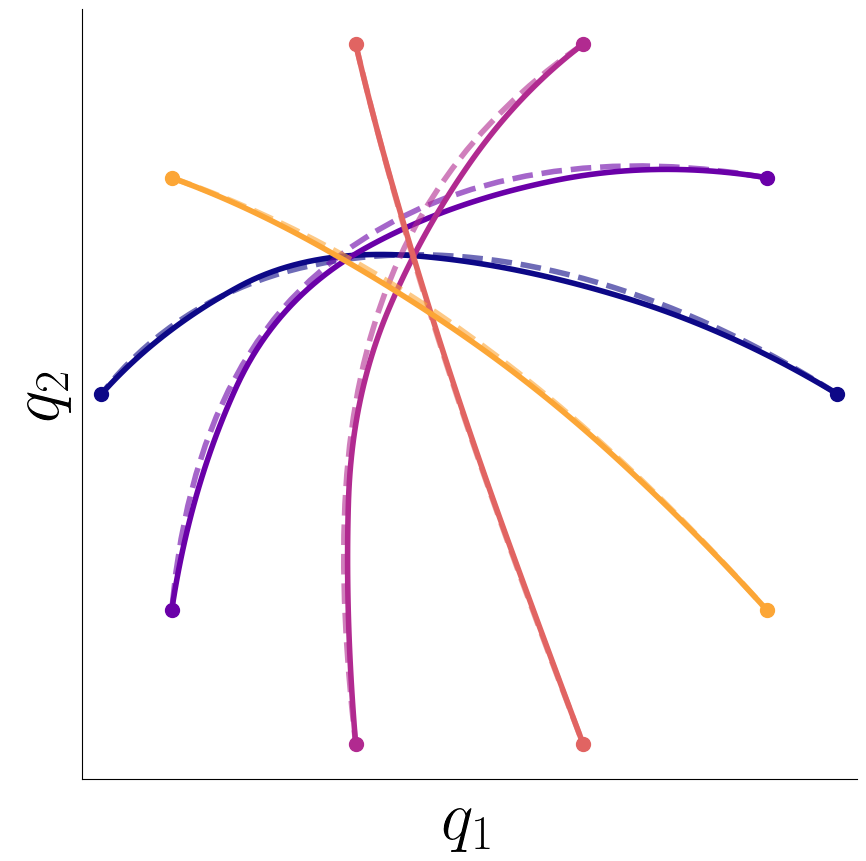}
        \caption{Kinetic-energy metric $\kineticmetric$}
        \label{fig:geods_inertia}
    \end{subfigure}
    \begin{subfigure}[b]{.45\linewidth}
        \includegraphics[width=.9\textwidth]{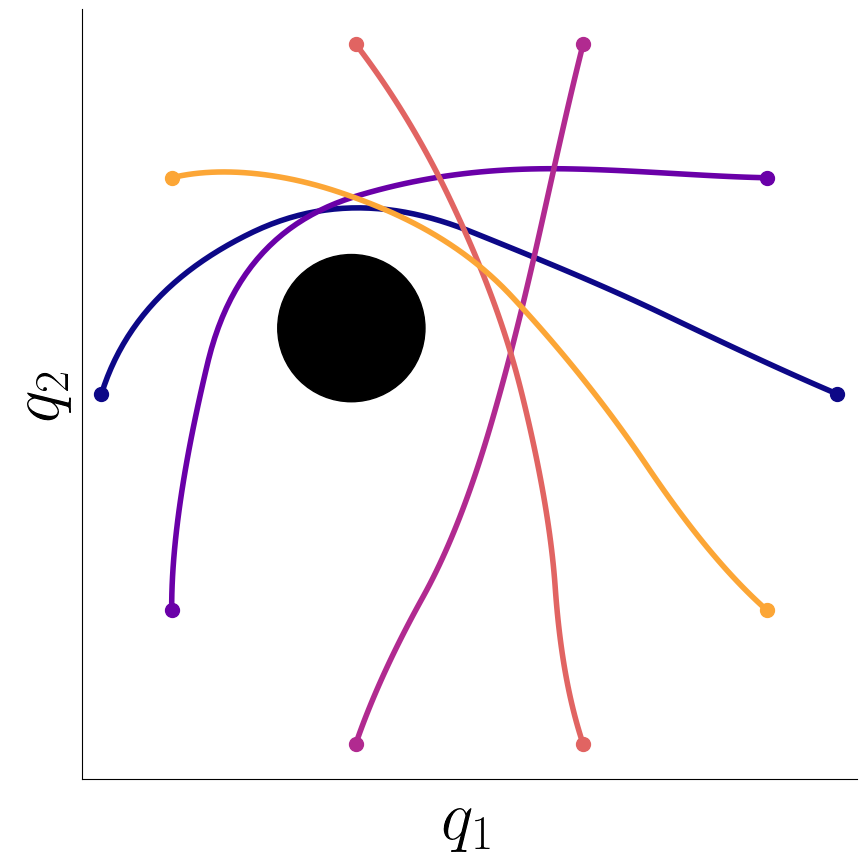}
        \caption{Collision-free metric $\metric$}
        \label{fig:geods_obstacle}
    \end{subfigure}
    \caption{Geodesics of a 2-DoFs planar robot. \emph{(a)} Geodesics approximated as cubic splines (\violetline) closely resembles solutions of the geodesic equation (\violetdashedline). \emph{(b)} Geodesics given by $\metric$ avoid the no-go region (\blackcircle), while maintaining the shape given by $\kineticmetric$.
    }
    \label{fig:geodesics}
    \vspace{-0.3cm}
\end{figure}

Given the collision-free metric defined in Eq.~\eqref{Eq:SumOfMetrics}, we aim at computing a geodesic connecting an initial to a final joint configuration. This corresponds to solving the geodesic equation~\eqref{Eq:Geodesic} with boundary conditions given by the initial and final configurations $\jointposition_i,\jointposition_f\in\configmanifold$. However, solving such boundary value problem for a system of coupled ODEs is notoriously hard. This is further exacerbated by the dimension of the robot configuration space, which is often $d\geq 6$ even for simple manipulators. Instead, we approximate geodesics $\jointposition(t)$ by cubic splines $\bm{\zeta}(t, \jointposition_c)$ with fixed boundary points given by the initial and final configurations $\jointposition_i,\jointposition_f$ and where $\jointposition_c=\{\jointposition_{c_k}\}_{k=1}^K$ defines $K$ control points $\jointposition_{c_k}\in\configmanifold$. These control points are optimized to minimize the (discretized) Riemannian kinetic energy~\eqref{Eq:KineticEnergy}
along $\bm{\zeta}$
\begin{equation}
    \jointposition_c = \argmin_{\jointposition_c} \int \frac{1}{2} \norm{\metric(\bm{\zeta}(t,\jointposition_c))}{\dot{\bm{\zeta}}(t,\jointposition_c)}^2 \mathrm{d} t, 
    \label{Eq:OptimizeGeodesicAsSpline}
\end{equation}
so that resulting splines resemble geodesics (see Figure~\ref{fig:geods_inertia}). 

This approach was also used in~\cite{BeikMohammadi22:ReactiveMotionGeneration} and is similar in spirit to the relaxation methods used to compute geodesics in~\cite{AlbuSchaffer22:ISRR,Laux21:GeodesicMotionPlanning}.
It is advantageous as it is guaranteed to find a trajectory between the given initial and final conditions. For instance, it avoids geodesics that stop in front of no-go regions due to local minima. Instead, the obtained trajectories wraps around obstacles, while being as close as possible to geodesics by minimizing the curve energy (see Figure~\ref{fig:geodesics}). In Section~\ref{sec:Experiments}, such geodesics are computed for $d=8$ in less than $0.5$s using Python code on a desktop with $3.70$GHz $\times24$ CPU and
$64$ GiB RAM. Notice that the kinetic energy of strict region-avoiding metrics $\metric$ goes to infinity in no-go regions, which are thus avoided by geodesics and spline approximations $\bm{\zeta}$ obtained via solutions of~\eqref{Eq:OptimizeGeodesicAsSpline}. 

\section{EXPERIMENTS}
\label{sec:Experiments}
We evaluate our approach on the humanoid robot \armarVI~\cite{Asfour2019:armar6}. We approximated the robot's collision geometry with capsules and boxes and used signed distance functions (SDF) to compute the direction from the closest point on the robot to a no-go region (leading to self-collision or to a collision with an obstacle) in task space. In the following, we used $\sigma=1$ for the inverse barrier, and fine-tuned the exponential barrier parameters for each experiment. 

\subsection{Joint Limits and Self-Collision Avoidance}

\begin{table}[t]
	\renewcommand*{\arraystretch}{1.2}
	\caption{Percentage of robot motions exceeding joint limits and resulting in self-collisions for different Riemannian metrics.}
	\label{Tab:EvaluationJointLimitsSelfCollisions}
	\begin{center}
    \vspace{-0.28cm}
		\begin{tabular}{c|c|c|c|}
		    & \multicolumn{3}{c|}{\% out of joint limits} \\
			& $\kineticmetric$ & $\metric$, $\barr_{\text{exp}}$ & $\metric$, $\barr_{\text{inv}}$ \\
			\hline
			$\mathsf{reaching}$ & $0$ & $0$ & $0$ \\
			$\mathsf{throwing}$ & $20.7$ & $20.3$ & $0$ \\ 
			$\mathsf{pointing}$ & $15.2$ & $15.8$ & $0$ \\ 
			$\mathsf{waving}$ (short) & $99.1$ & $99.1$ & $0$ \\
			$\mathsf{waving}$ (long) & $46.8$ & $47.6$ & $0$ \\ 
            $\mathsf{arms~crossing}$ & $100$ & $100$ & $0$ \\
			\hline
            \hline
            & \multicolumn{3}{c|}{\% with self-collisions} \\
            & $\kineticmetric$ & $\metric$, $\barr_{\text{exp}}$ & $\metric$, $\barr_{\text{inv}}$ \\
            \hline
            $\mathsf{reaching}$ & $0$ & $0$ & $0$  \\
			$\mathsf{throwing}$& $2.9$ & $3.2$ & $0$\\ 
			$\mathsf{pointing}$ & $0$ & $2.3$ & $0$ \\ 
			$\mathsf{waving}$ (short)  & $24.7$ & $19.9$ & $0$ \\
			$\mathsf{waving}$ (long) & $0$ & $0$ & $0$\\ 
            $\mathsf{arms~crossing}$ & $54$ & $65$ & $0$ \\
			\hline
		\end{tabular}
	\end{center}
\vspace{-0.55cm}
\end{table}

We first evaluate our approach to avoid the robot's joint limits and self-collisions during motion generation. 
We consider robot motions obtained by transferring human movements described as sequences of geodesics in the human configuration space to the robot configuration space. Here, we use $5$ human movements from the KIT whole-body human motion database~\cite{Mandery2016b:MotionDatabase}\footnote{\url{https://motion-database.humanoids.kit.edu/}}, namely  one $\mathsf{reaching}$, one $\mathsf{throwing}$, one $\mathsf{pointing}$ (successively up, horizontally, and down), and two $\mathsf{waving}$ (in greeting) motions\footnote{Motions identified in the database as take\_book\_from\_shelf\_right\_arm\_01, throw\_right01, point\_at\_right03, wave\_left01, and waving\_neutral04.}. As all these movements mostly involve the joints of one arm, we restrict the transfer to \armarVI's $8$-DoFs left or right arm. 
We transfer these motions to \armarVI by leveraging the Riemannian transfer framework introduced in our previous work~\cite{Klein2022}. Namely, each motion is first mapped onto the Master Motor Map (MMM) model~\cite{Mandery2016b:MotionDatabase}, which is a reference model of the human body, and segmented into a sequence of geodesics using Riemannian motion segmentation. The key points of the motion, given by hand pose at the start and the end of each geodesic, are mapped from the human task space to the robot task space. In practice, the hand position is scaled proportionally to the total arm length.
Each extracted human geodesic is then reproduced as a geodesic in the robot configuration manifold.
Given the desired initial and final hand poses $\bm{x}_i^{g},\bm{x}_f^{g}$ of the $g$-th geodesic, we aim at finding a geodesic $\jointposition(t)$ in the robot configuration manifold $\configmanifold$ whose boundary conditions $\jointposition_i^g,\jointposition_f^g\in\configmanifold$ satisfy $f(\jointposition_i^g) = \bm{x}_i^{g}$ and $f(\jointposition_f^g)=\bm{x}_f^{g}$.
As the geodesics are followed sequentially, we assume that the initial condition $f(\jointposition_i^g) = \bm{x}_i^{g}$ is already satisfied. Therefore, we find a final configuration satisfying $f(\jointposition_f^g)=\bm{x}_f^{g}$ using, e.g., IK, and compute the corresponding geodesic as described in Section~\ref{sec:GeodesicMotionGeneration}.
Following this approach, each transferred geodesic constitutes a minimum-energy trajectory accounting for the robot's own inertial properties, while the overall sequence conserves the main characteristics of the original human motion.

\begin{figure}
    \centering
    \begin{subfigure}[b]{.46\linewidth}
        \includegraphics[width=\textwidth]{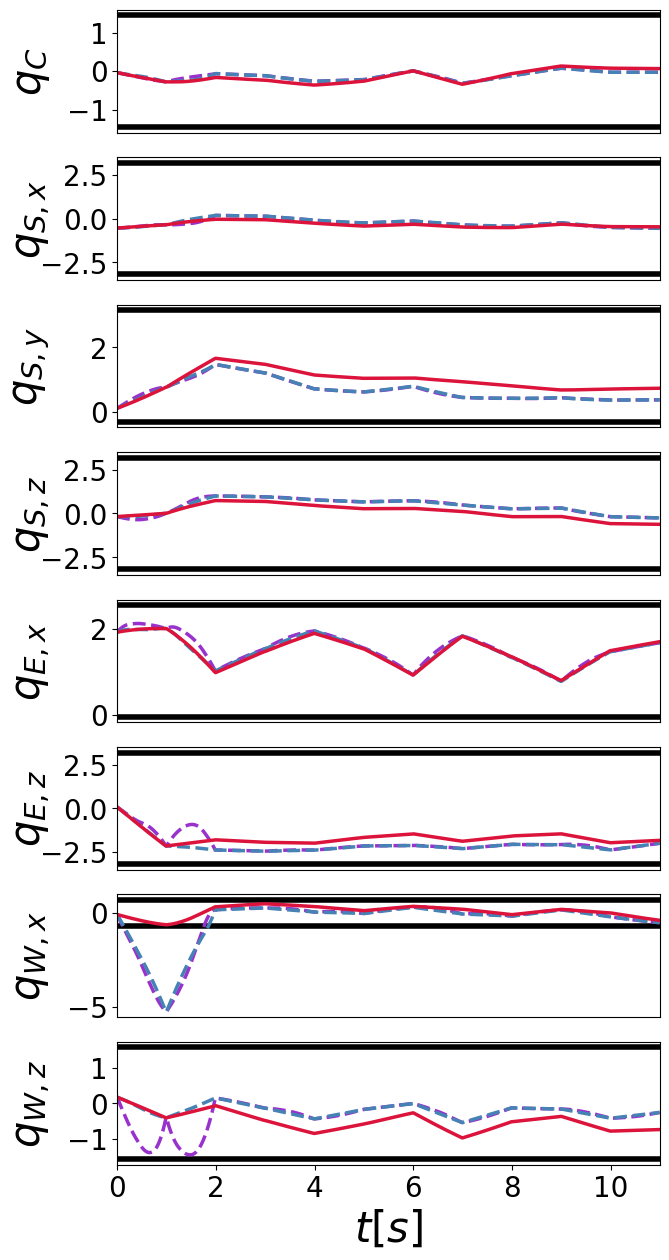}
        \caption{$\mathsf{Pointing}$ motion}
        \label{fig:jointlimits_pointing}
    \end{subfigure}
    \begin{subfigure}[b]{.46\linewidth}
        \includegraphics[width=\textwidth]{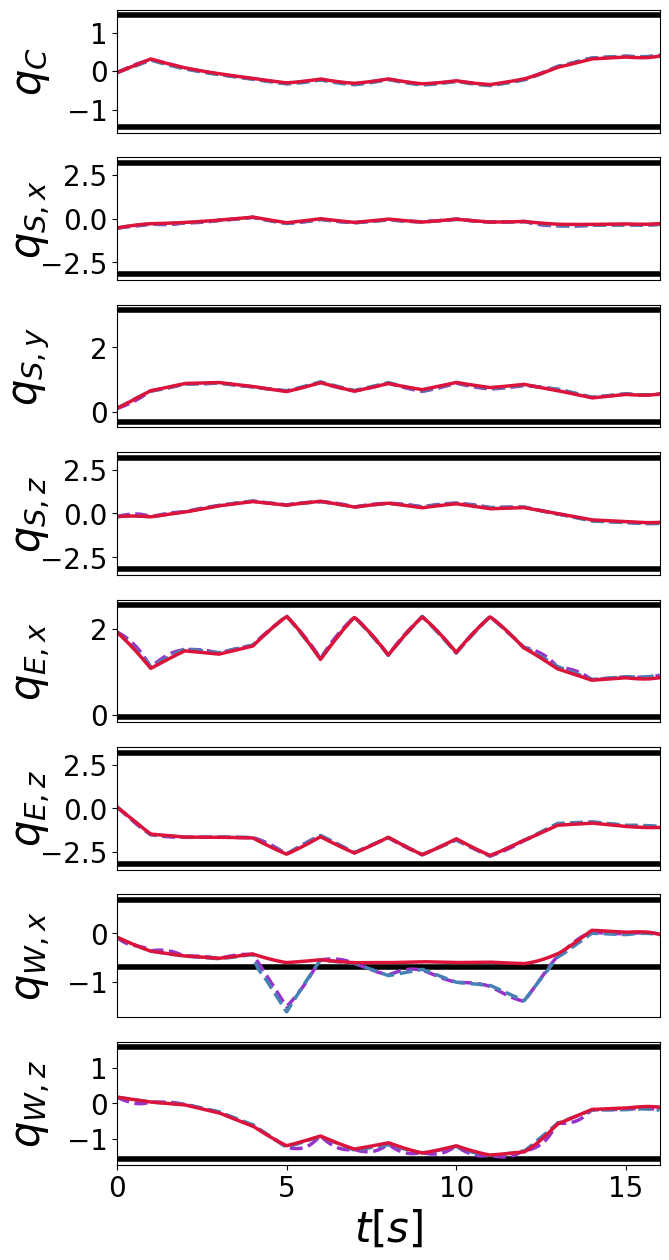}
        \caption{$\mathsf{Waving}$ (long) motion}
        \label{fig:geods_obstacle}
    \end{subfigure}
    \caption{Transfer of human motions to \armarVI. The trajectories of the clavicle ($q_C$), shoulder ($q_{S,\cdot}$), elbow ($q_{E,\cdot}$), and wrist ($q_{W,\cdot}$) joints are depicted along with joint limits (\blackline). The trajectories are obtained using the Riemannian model with the kinetic-energy metric (\violetdashedline), and the collision-free metric based on the exponential (\bluedashedline) and on the inverse (\redline) barriers.
    }
    \label{fig:Jointlimits}
    \vspace{-0.6cm}
\end{figure}

We compare the transferred robot motions obtained by endowing the robot configuration space with the kinetic-energy metric $\kineticmetric$, and the collision-free metric $\metric$~\eqref{Eq:CollisionFreeMetric} using region-avoiding metrics based on the exponential and inverse barriers $\barr_{\text{exp}}$ and $\barr_{\text{inv}}$. As shown in Table~\ref{Tab:EvaluationJointLimitsSelfCollisions}, the motions generated using the kinetic-energy metric exceed joint limits and results in self-collisions along the trajectory. Note that this is the expected behavior, as neither joint limits nor collisions are considered by this metric. In contrast, the proposed inverse-barrier metric reliably generates feasible robot motions. Moreover, we observe that the exponential-barrier metric barely reduces the proportion of joint limits excesses and of self-collisions. As explained in Section~\ref{sec:RiemannianMetricsRegionAvoidance}, this is due to the fact that exponential barriers do not reach infinite values in no-go regions, and thus cannot guarantee that they are strictly avoided by geodesics. Importantly, similar behaviors are observed for inverse- and exponential-barrier metrics independently of the values of the parameters $\sigma$ and $\lambda$. Figure~\ref{fig:Jointlimits} shows the trajectories of the $8$ joints of the right arm. We observe that trajectories generated by both collision-free metrics closely resemble those generated by the kinetic-energy metric. In other words, the inverse barriers modify the kinetic-energy metric so that geodesics stay within the robot's joint limits but are not influenced away from no-go regions. This is particularly visible for the wrist joint $q_{W,x}$. Joint limits are not respected by the exponential barriers. Transitions between geodesics may lead to sudden accelerations which can be smoothed out by encoding transitions with larger numbers of smaller geodesics~\cite{Klein2022}. 

\begin{figure}
    \centering
    \includegraphics[trim={15cm 3.5cm 10cm 0cm},clip,width=.115\textwidth]{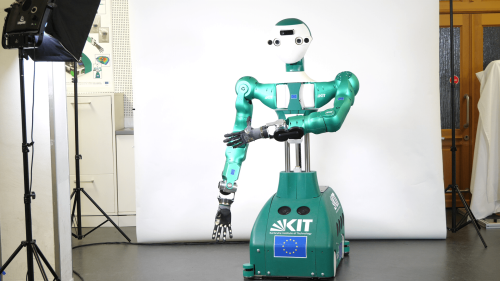}
    \includegraphics[trim={15cm 3.5cm 10cm 0cm},clip,width=.115\textwidth]{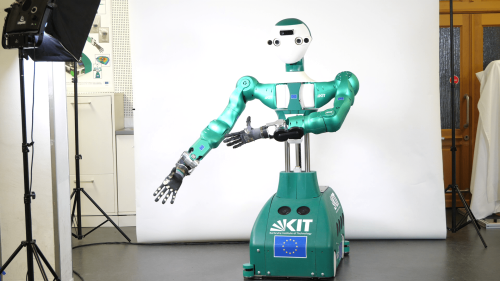}
    \includegraphics[trim={15cm 3.5cm 10cm 0cm},clip,width=.115\textwidth]{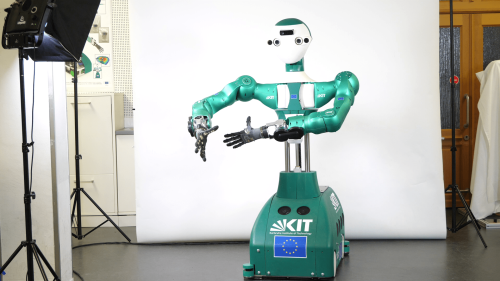}
    \includegraphics[trim={15cm 3.5cm 10cm 0cm},clip,width=.115\textwidth]{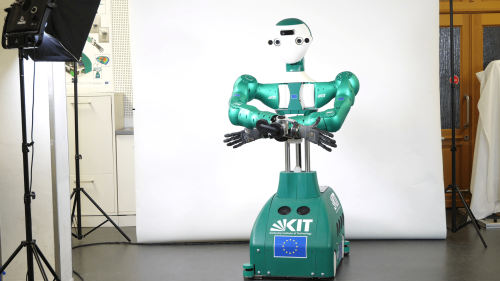}
    \caption{$\mathsf{Arms~crossing}$ motion realized by following a geodesic generated from the inverse-barrier-based collision-free metric $\metric$.}
    \label{fig:Selfcollision}
    \vspace{-0.4cm}
\end{figure}

As most transferred motions only result in few self-collisions, including when generated solely with the kinetic-energy metric, we evaluate our approach in a task that explicitly requires efficient self-collision avoidance. To do so, we consider an $\mathsf{arms~crossing}$ motion, where the robot follows a single geodesic to cross its right arm over the left one. Figure~\ref{fig:Selfcollision} shows snapshots of the motion executed using the collision-free metric $\metric$ computed using inverse barriers. We observe that the resulting trajectory successfully avoids collisions between the two robot arms. In particular, the clavicle and elbow joints are leveraged so that the right arm rotates around the left hand before reaching the final crossed-arm posture. As shown in Table~\ref{Tab:EvaluationJointLimitsSelfCollisions}, geodesics generated by the kinetic-energy and exponential-barrier metrics result in a high proportion of collisions between both robot arms.

\begin{figure}
    \centering
    \begin{subfigure}[b]{.245\linewidth}
        \includegraphics[trim={4cm 7cm 11cm 6cm},clip,width=\textwidth]{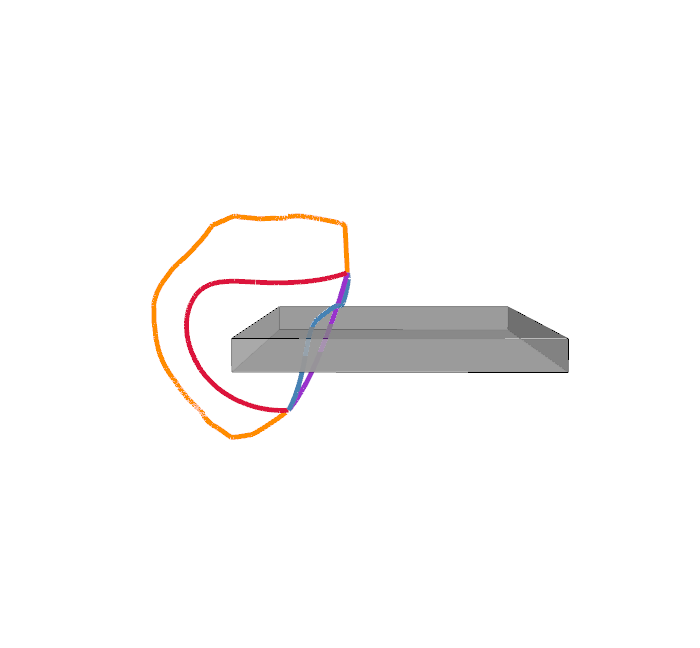}
        \caption{Hand motions.}
        \label{fig:table_trajs}
    \end{subfigure}
    \begin{subfigure}[b]{.74\linewidth}
       \includegraphics[trim={15cm 3cm 13cm 0cm},clip,width=.32\textwidth]{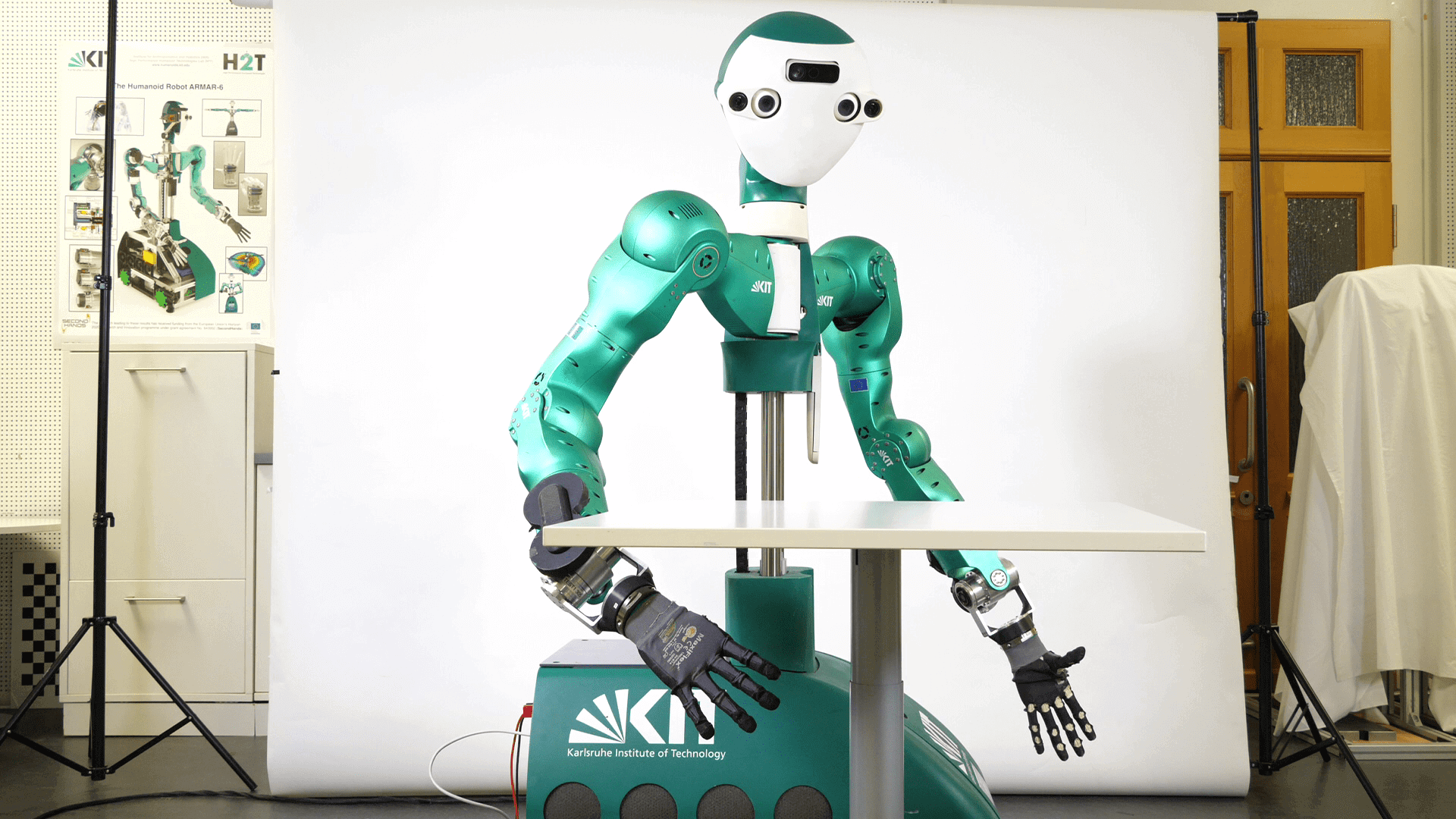}
        \includegraphics[trim={15cm 3cm 13cm 0cm},clip,width=.32\textwidth]{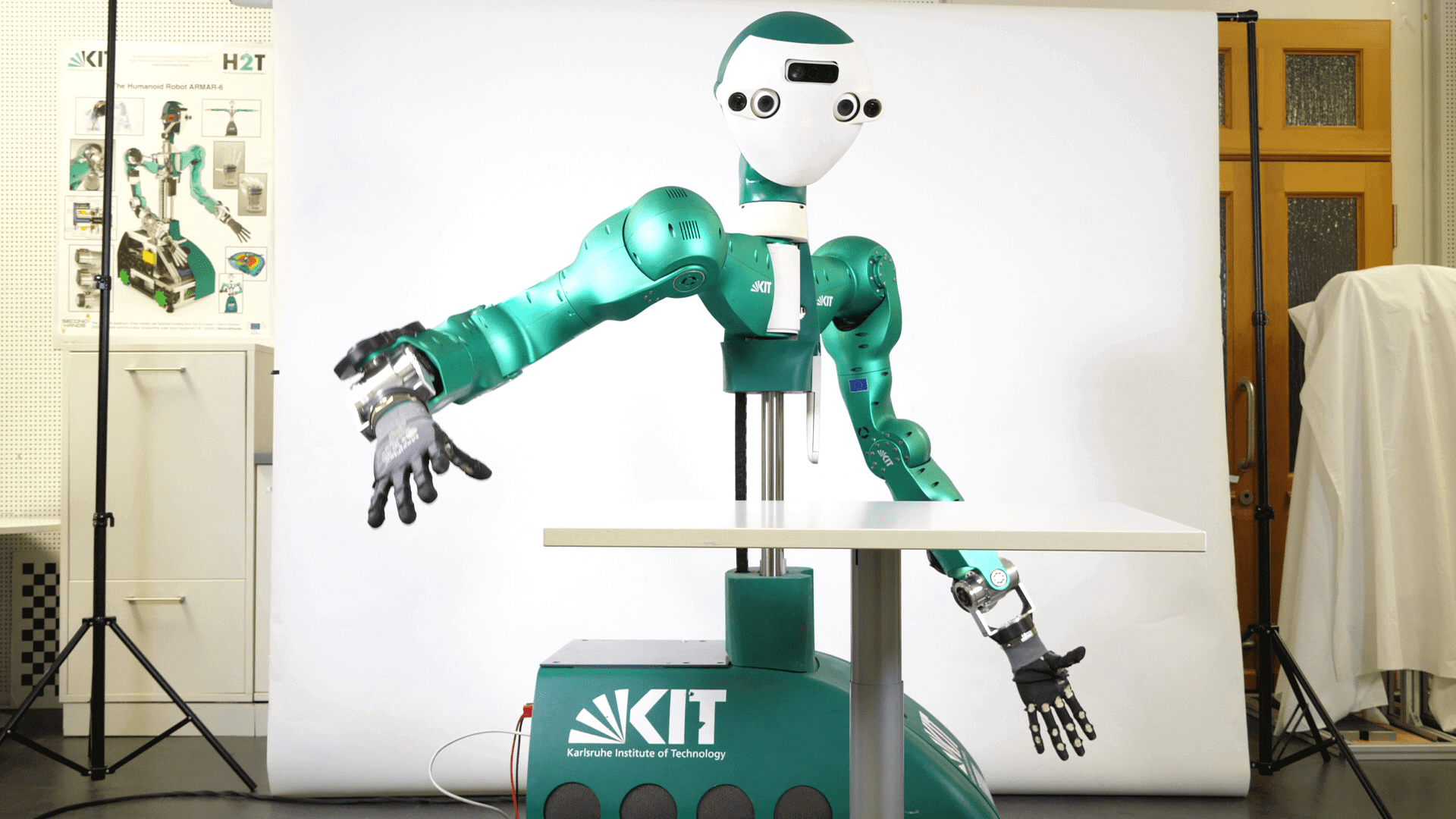}
        \includegraphics[trim={15cm 3cm 13cm 0cm},clip,width=.32\textwidth]{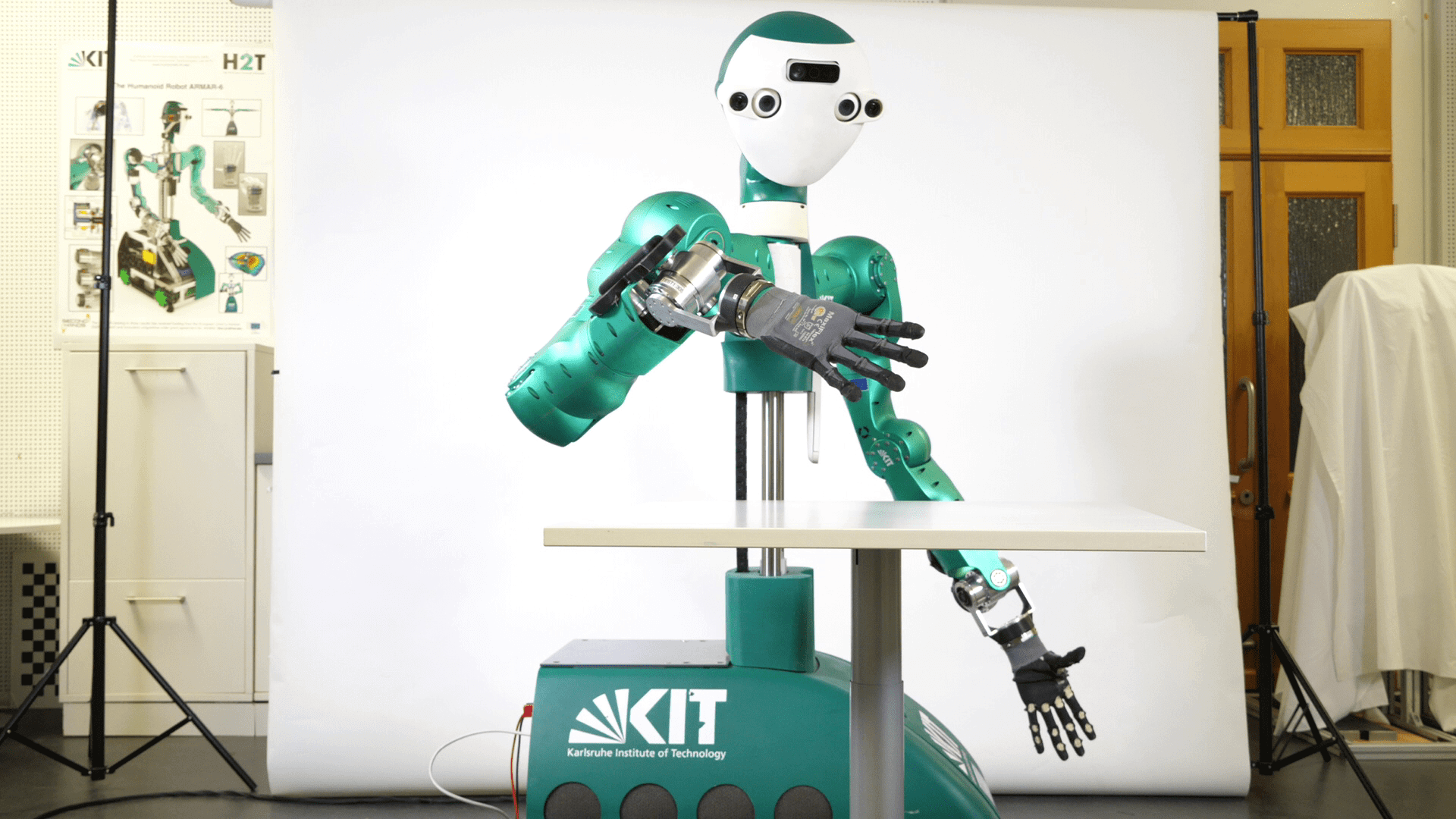}
        \caption{Inverse-barrier-based collision-free geodesic.}
        \label{fig:table_inv}
    \end{subfigure}
    \caption{Geodesic trajectories generated by different Riemannian metric to avoid a table. \emph{(a)} The hand trajectories obtained following geodesics using the kinetic-energy metric (\violetline), and the collision-free metric based on the exponential (\blueline) and on the inverse (\redline) barriers. The trajectory obtained with CollisionIK is displayed as a baseline (\orangeline). \emph{(b)} Snapshots of the resulting robot motion.}
    \label{fig:TableAvoidance}
    \vspace{-0.5cm}
\end{figure}

\subsection{Obstacle Avoidance}
Next, we evaluate our approach to generate trajectories avoiding obstacles in task space. To do so, we consider a scenario where the robot's hand is initially placed below a table and should be placed at a final location above the table. The corresponding motion is achieved by following a single geodesic. We compare the geodesics obtained by endowing the robot configuration space with the kinetic-energy metric $\kineticmetric$, and the collision-free metric $\metric$~\eqref{Eq:CollisionFreeMetric} using region-avoiding metrics based on the exponential and inverse barriers $\barr_{\text{exp}}$ and $\barr_{\text{inv}}$. As shown in Table~\ref{Tab:EvaluationObstacleAvoidance} and in Figure~\ref{fig:table_trajs}, the geodesic generated by the kinetic-energy metric passes through the table, thus resulting in a high proportion of collisions. This is expected, as obstacles are not considered by the kinetic-energy metric. As for joint limits and self-collision avoidance, the exponential-barrier metric does not avoid the obstacle. This behavior occurs independently of the values of the barrier parameters. As shown in Figure~\ref{fig:table_trajs}, the resulting geodesic slightly diverges from the kinetic-energy geodesic and tends towards the border of the table. However, it does not suffice to avoid the collision. Instead, the inverse-barrier metric generates a geodesic around the table, successfully avoiding collisions. Table~\ref{Tab:EvaluationObstacleAvoidance}-\emph{bottom} compares the Riemannian length of the obtained trajectories. The lengths are all computed with respect to the kinetic-energy metric. As a baseline, we also consider a trajectory generated using CollisionIK~\cite{Rakita21:CollisionIK} for which the final hand pose was given as goal pose and whose cost weights were adapted to our scenario. The shortest trajectory was generated by the kinetic inertia metric $\kineticmetric$. This is expected, as it corresponds to a geodesic, i.e., a shortest path, in the corresponding manifold. Moreover, the trajectory generated by the inverse-barrier metric is shorter than the trajectory generated by CollisionIK, i.e., closer to a geodesic with respect to $\kineticmetric$. 

\begin{table}[t]
	\renewcommand*{\arraystretch}{1.2}
	\caption{Percentage of robot motions colliding with the obstacle and motion length for different Riemannian metrics. Motion lengths are computed with respect to the kinetic-energy metric.}
	\label{Tab:EvaluationObstacleAvoidance}
    \vspace{-0.3cm}
	\begin{center}
		\begin{tabular}{c|c|c|c|c|} 
		    & \multicolumn{4}{c|}{\% with obs. collision} \\
			& $\kineticmetric$ & $\metric$, $\barr_{\text{exp}}$ & $\metric$, $\barr_{\text{inv}}$ & C-IK \\
			\hline
			$\mathsf{table}$ & $64$ & $50$ & $0$ & $0$ \\
			$\mathsf{waving}$ (long) with obs.& $35.2$& $37.8$& $0$ & $0$ \\ 
			\hline
            \hline
            & \multicolumn{4}{c|}{Riemannian length $\ell_{\kineticmetric}$} \\
            & $\kineticmetric$ & $\metric$, $\barr_{\text{exp}}$ & $\metric$, $\barr_{\text{inv}}$ & C-IK \\
            \hline
            $\mathsf{table}$ & $1.20$ & $1.40$ & $3.01$ & $6.14$\\
			$\mathsf{waving}$ (long) with obs. & $12.54$ & $13.47$ & $21.35$ & N/A \\ 
            \hline
		\end{tabular}
	\end{center}
 \vspace{-0.65cm}
\end{table}

\begin{figure}
    \centering
    \begin{subfigure}[b]{.245\linewidth}
        \includegraphics[trim={6.5cm 10cm 7.5cm 3.2cm},clip,width=\textwidth]{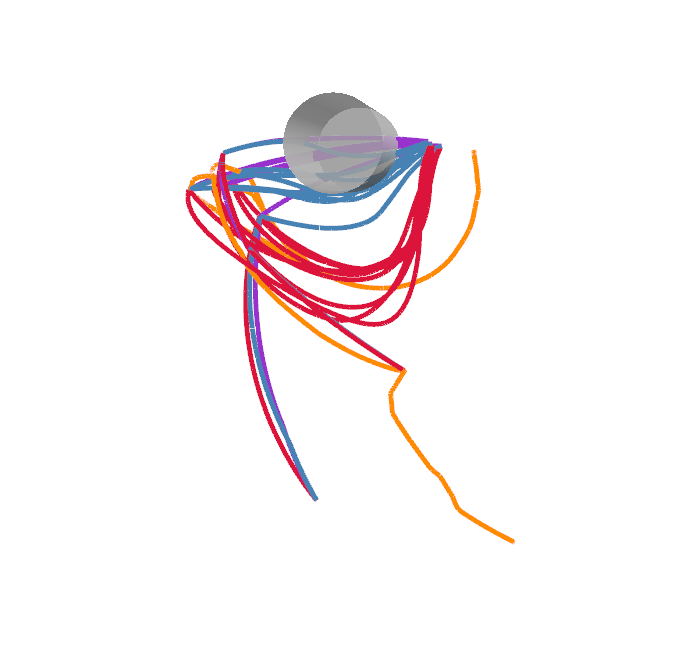}
        \caption{Hand motions.}
        \label{fig:wavingobs_trajs}
    \end{subfigure}
    \begin{subfigure}[b]{.74\linewidth}
       \includegraphics[trim={26cm 17cm 20cm 1cm},clip,width=.32\textwidth]{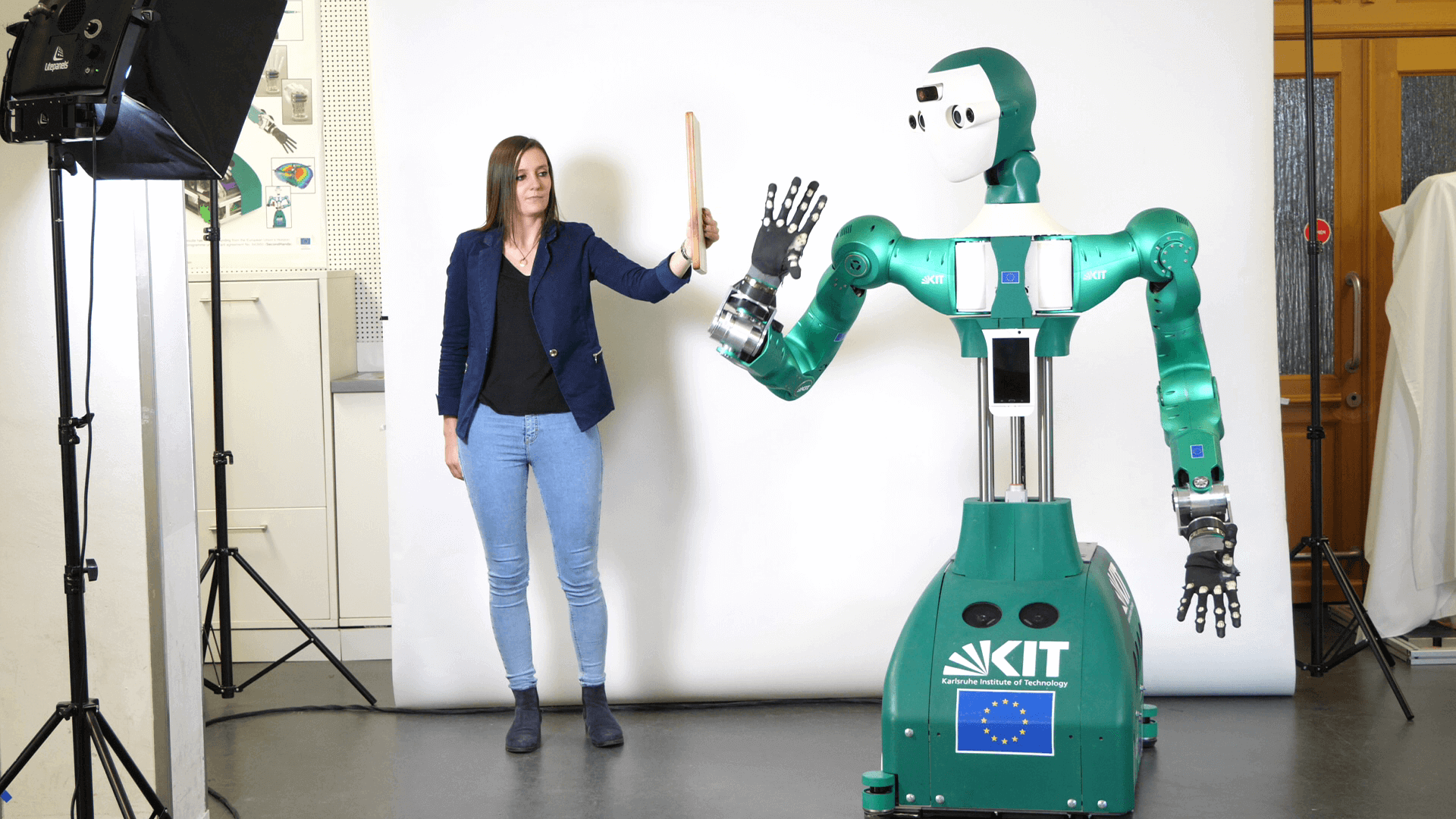}
        \includegraphics[trim={26cm 17cm 20cm 1cm},clip,width=.32\textwidth]{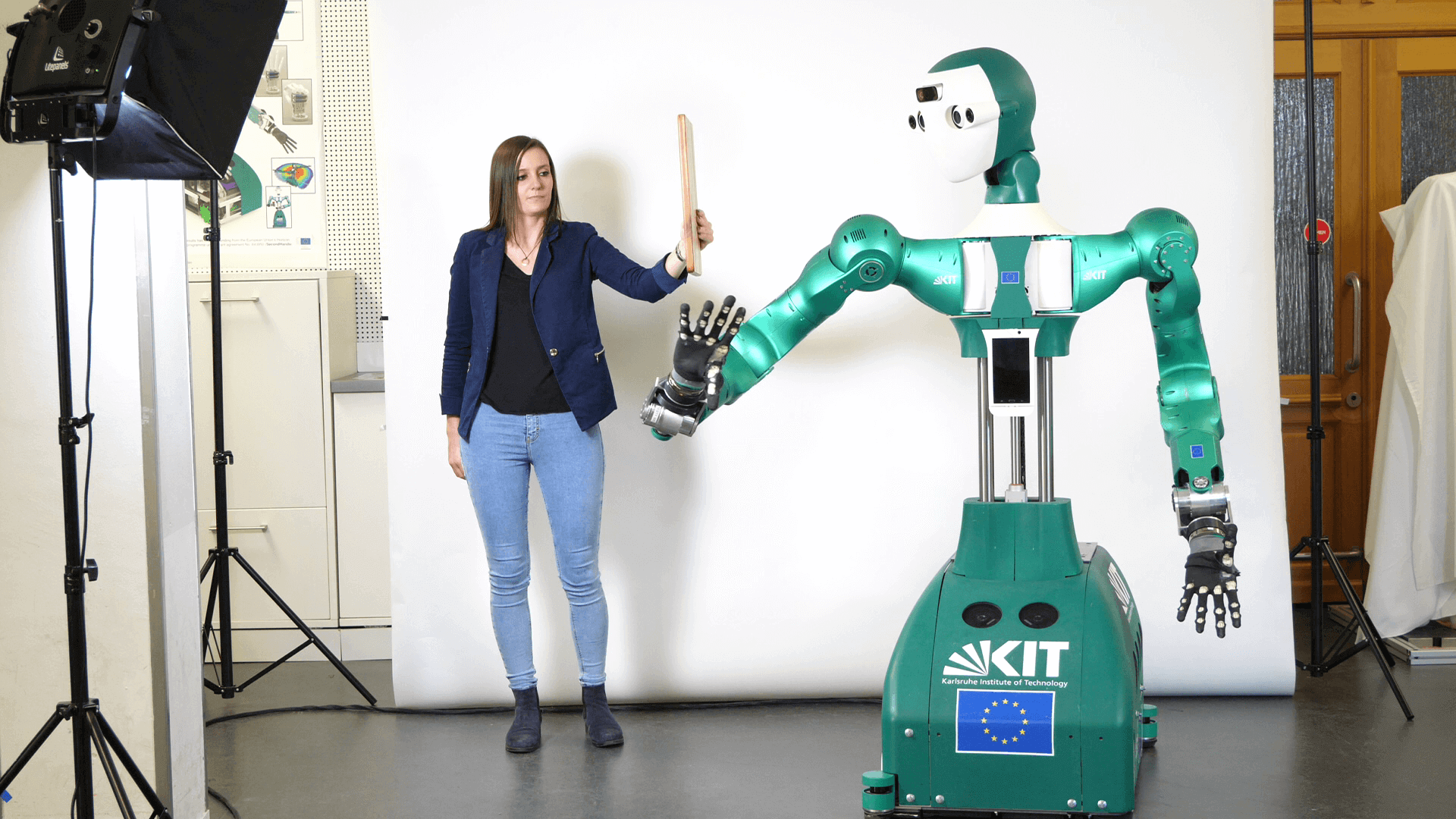}
        \includegraphics[trim={26cm 17cm 20cm 1cm},clip,width=.32\textwidth]{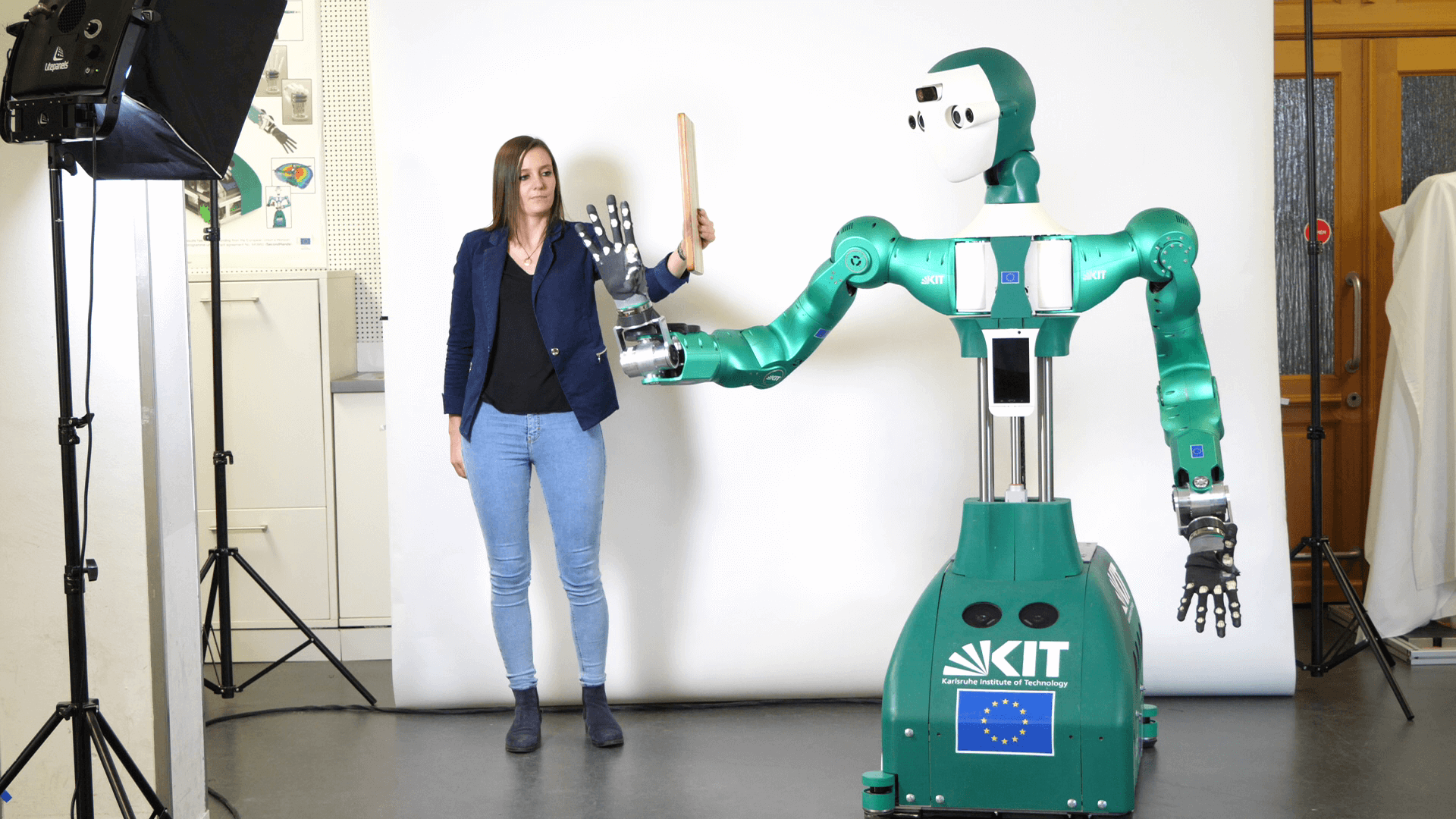}
        \caption{Inverse-barrier-based collision-free trajectory.}
        \label{fig:wavingobs_inv}
    \end{subfigure}
    \caption{Sequences of geodesics generated by different Riemannian metrics to avoid an obstacle during $\mathsf{waving}$. \emph{(a)} Hand trajectories obtained by following geodesics using the kinetic-energy metric (\violetline), and the collision-free metric with exponential (\blueline) and inverse (\redline) barriers. The CollisionIK trajectory is displayed as a baseline (\orangeline). \emph{(b)} Snapshots of the resulting robot motion.}
    \label{fig:WavingObstacleAvoidance}
    \vspace{-0.6cm}
\end{figure}

To evaluate the ability of our proposed metric to avoid obstacles along longer trajectories, we then consider the long $\mathsf{waving}$ motion transferred in the previous section augmented with an obstacle. As in the previous section, the trajectories are generated as sequences of geodesics. As shown in Figure~\ref{fig:WavingObstacleAvoidance} and Table~\ref{Tab:EvaluationObstacleAvoidance}, the geodesics generated by the inverse-barrier metric successfully avoid the obstacle, while maintaining the general shape of the waving motion. Interestingly, part of the geodesics generated by the exponential-barrier metric successfully avoid the obstacle (see Figure~\ref{fig:wavingobs_trajs}). This confirms that the exponential barrier encourages geodesics to avoid no-go regions. However, as previously mentioned, strict avoidance is not guaranteed. 
As for the $\mathsf{around~table}$ motion, the shortest trajectory is generated by the kinetic-energy metric. Although the inverse-barrier metric generates longer trajectories with respect to the kinetic-energy metric, its kinetic-energy metric component encourages trajectories with low energy with respect to the kinetic-energy metric. This results in trajectories passing below the obstacle to avoid it. 
CollisionIK was prone to local minima in this scenario and did not always find a way from one side of the obstacle to the other (see Figure~\ref{fig:wavingobs_trajs}). Therefore, in contrast to our approach, it could not achieve the entire waving motion.

\section{CONCLUSION}
\label{sec:Conclusion}
This paper studied region-avoiding metrics allowing the generation collision-free robot motions as geodesics on a Riemannian manifold. Specifically, we proposed to modify the Riemannian metric of the configuration space manifold to incorporate additional metric terms guaranteeing joint limits, self-, and obstacle avoidance. To do so, we introduced a class of Riemannian metrics based on barrier functions and derived a sufficient and necessary condition which guarantees that geodesics will avoid given regions. This condition is satisfied by inverse barriers, which guarantee strict region avoidance as opposed to exponential barriers. We showed how these metrics are adapted to consistently generate geodesic accelerations away from no-go regions, while having no influence on geodesics traveling along these regions.

Our experiments showed that the proposed inverse-barrier Riemannian metric generates geodesics corresponding to collision-free trajectories within joint limits. Moreover, the resulting trajectories are close to geodesics with respect to the kinetic-energy metric, and thus result in dynamic-aware and energy-efficient robot motions. Importantly, the proposed region-avoiding metrics are not limited to the configuration space manifold, but can, in principle, be applied to any other Riemannian manifold. Moreover, other barrier functions, e.g., with compact support, satisfying the derived sufficient and necessary condition may also be considered.
Although our metrics generate geodesic acceleration depending on the direction to a no-go region, they do not differentiate between geodesics approaching or going away from the given region. This limitation comes from the fact that Riemannian metrics are functions of positions on the manifold, while differentiating between these two types of trajectories requires taking velocities into account. Alternative metrics considering velocities will be explored as future work.

\bibliographystyle{IEEEtran}
\vspace{-0.2cm}
\bibliography{References} 

\end{document}